\documentclass{article}

\usepackage{arxiv}
\usepackage[utf8]{inputenc} % allow utf-8 input
\usepackage[T1]{fontenc}    % use 8-bit T1 fonts
\usepackage{hyperref}       % hyperlinks
\usepackage{url}            % simple URL typesetting
\usepackage{booktabs}       % professional-quality tables
\usepackage{amsfonts}       % blackboard math symbols
\usepackage{nicefrac}       % compact symbols for 1/2, etc.
\usepackage{microtype}      % microtypography
\usepackage{lipsum}		% Can be removed after putting your text content
\usepackage{graphicx}
\usepackage{natbib}
\usepackage{doi}
\usepackage{multirow}
\usepackage{amsmath}
\usepackage{comment}

\title{Design strategies for controlling neuron-connected robots using reinforcement learning}

\date{} 					% Or removing it

\author{ 
    {Haruto Sawada} \\
    Department of Mechano Informatics \\
    The University of Tokyo, Tokyo, Japan \\
    \AND
    {Naoki Wake}\\
    Applied Robotics \\
    Microsoft, Redmond, WA 98052, USA \\
	\texttt{naoki.wake@microsoft.com} \\
	\And
	{Kazuhiro Sasabuchi}\\
    Applied Robotics \\
    Microsoft, Redmond, WA 98052, USA \\
	\AND
	{Jun Takamatsu} \\
    Applied Robotics \\
    Microsoft, Redmond, WA 98052, USA \\
    \AND
	{Hirokazu Takahashi} \\
    Department of Mechano Informatics \\
    The University of Tokyo, Tokyo, Japan \\
    \AND
	{Katsushi Ikeuchi} \\
    Applied Robotics \\
    Microsoft, Redmond, WA 98052, USA \\
}

\begin{document}
\maketitle
\begin{abstract}
Despite the growing interest in robot control utilizing the computation of biological neurons, context-dependent behavior by neuron-connected robots remains a challenge. Context-dependent behavior here is defined as behavior that is not the result of a simple sensory-motor coupling, but rather based on an understanding of the task goal. This paper proposes design principles for training neuron-connected robots based on task goals to achieve context-dependent behavior. First, we employ deep reinforcement learning (RL) to enable training that accounts for goal achievements. Second, we propose a neuron simulator as a probability distribution based on recorded neural data, aiming to represent physiologically valid neural dynamics while avoiding complex modeling with high computational costs. Furthermore, we propose to update the simulators during the training to bridge the gap between the simulation and the real settings. The experiments showed that the robot gradually learned context-dependent behaviors in pole balancing and robot navigation tasks. Moreover, the learned policies were valid for neural simulators based on novel neural data, and the task performance increased by updating the simulators during training. These results suggest the effectiveness of the proposed design principle for the context-dependent behavior of neuron-connected robots.
\end{abstract}

% keywords can be removed

\section{Introduction}
While living organisms can learn behaviors appropriate to a given context through the function of neurons, there is no generic methodology for robots to autonomously acquire context-dependent behaviors. We define context-dependent behavior as adaptive behavior based on the understanding of the task goal, beyond a simple sensory-motor coupling. Robots may be able to learn context-dependent behaviors by harnessing the computational power of biological neurons. A typical framework for robot control utilizing biological nervous systems is Animat, which operates the robot based on the activity of dissociated neuronal culture evoked by electrical stimuli that reflect the robot's sensory input (e.g., \cite{demarse2001neurally, chao2008shaping, bakkum2008spatio, yada2021physical}). These studies have demonstrated the usefulness of neural computation focusing on ``Braitenberg vehicle-type'' robot control, in which tasks are accomplished through simple sensory-motor coupling. Although the realization of context-dependent behavior by Animat is expected to provide one methodology for intelligent control of robots, to the best of our knowledge, no study has successfully learned context-dependent behavior. Context-dependent behavior requires the system to recognize the task goal, but it is generally challenging to represent a task goal in a biological neural network with complex and fluctuating dynamics.

This study aims to achieve context-dependent behavior in Animats by addressing three challenges associated with learning Animat based on task goals. First is a methodology for having the system represent task goals. The second is a methodology to deal with the dynamics of neurons. The third is a methodology to handle the change of neural responsiveness caused by network plasticity and cellular damage. Here, we propose a design principle to address these issues. First, we propose using reinforcement learning (RL) and a neuron simulator for training Animats in the light of task goal. Second, we propose a neuron simulator based on data recorded from biological neurons instead of mathematical models. Finally, we propose to keep the neuron simulator ``fresh'' during RL by periodically updating the simulators. We refer to this technique as \textit{parameter shadowing}.

We tested the usefulness of these design principles through experiment of in which Animats learned context-dependent behaviors. The experiments showed that the Animats gradually learned context-dependent behaviors for pole balancing and robot navigation in the simulation. Moreover, the learned policies were valid for neural simulators based on novel neural data, and the performance of robots was improved by updating the simulators during training. These results suggest the effectiveness of the proposed design principle for the context-dependent behavior of neuron-connected robots. The main contributions of this paper are the proposal of design principles for applying RL to Animat and the demonstration of context-dependent behavior in Animat.

In what follows, Section 2 describes the design principles in detail, showing the position of this paper in relation to previous studies. Section 3 describes the implementation of Animats in this study based on the design principles. In Section 4, we conduct experiments using biological neurons and the implemented Animats, and show that Animat's context-dependent behavior can be efficiently learned through RL and parameter shadowing. We discuss the experimental results in Section 5 and conclude in Section 6.

\section{Design principles for controlling neuron-connected robots}\label{design}
We explain three design principles to address difficulties for achieving context-dependent behavior by neuron-connected robots. The first challenge is how to adjust the control loop to achieve context-dependent behavior. We propose using RL to learn how to stimulate neurons based on a reward function that reflects the optimality of behaviors. The second challenge is to model the neural response to electrical stimulation. Focusing on the resonant nature of the neurons, we propose to model the relationship between stimulus frequency and the expected total number of firings elicited from the stimulus. The third challenge is how to minimize the effect of changes in neural responsiveness. We propose to update the simulator based on measured neuronal data during training.

\subsection{Design for learning context-dependent behavior}

Robot control research has long used reinforcement learning (RL) to learn optimal behavior in a given environment. Here, we propose to use RL to learn context-dependent behaviors of Animats. By defining a reward function that reflects the optimality of the behavior, we hypothesized that Animat could learn how to stimulate neurons to achieve its optimal behavior. While there are several studies that applied RL to biological neurons\cite{wulfing2019adaptive}, their goal was to control the neural activities rather than a embodied robots. To the best of our knowledge, there was no study that applied RL in neuron-connected robots.

\subsection{Design for modeling neural response to electrical stimulation}\label{neuron_model}
Because robot control using RL requires many iterations to learn the policy, simulation is generally utilized instead of real setups. As for the RL for Animat control, the use of actual equipment also entails various difficulties. First, the use of real Animats carries the risk of neural fatigue and cell death associated with a number of electrical stimuli. Second, the parallel computation cannot be used for the RL because of the difficulty of preparing a physical culture dish with identical dynamics. To efficiently perform RL for Animat, it is necessary to prepare a neuron simulator that simulates neuronal dynamics. 

However, the complex interactions inside the network make it difficult to simulate the full dynamics of cultured neurons. To solve this problem, we propose to use neural response data to electrical stimuli instead of mathematical models such as spiking neural networks. If a stimulus is defined for which a certain reproducibility can be assumed for the neural response, a data-based neuron simulator would predict the neural response more simply and at a lower computational cost than using mathematical models.

As an example of such a neuron simulator, this study focused on the resonant nature of the neurons based on their electrical characteristics ~\cite{buzsaki2004neuronal, thomson2003interlaminar} and defined the stimulus frequency as the input and the total number of evoked firings as the output. The simulator was prepared as a probability distribution of the output values against the stimulus frequency based on actual recordings.

\subsection{Design to minimize the effect of changes in neural responsiveness}\label{parameter_shadowing}
No matter what model is prepared, difficulties remain in reflecting the evolution of neural dynamics in multiple time scales. Temporal evolution results in a discrepancy between the actual neurons and simulators, and could limit the performance when connecting a policy to a system with real neurons. This is a form of sim-to-real problem, which is a common problem in RL.

\begin{figure}[ht]
	\centering
	\includegraphics[width=0.6\linewidth]{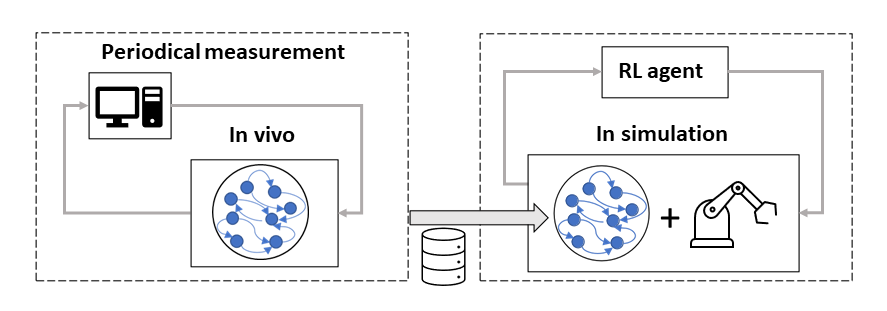}
	\caption{Paradigm of parameter shadowing.}
	\label{fig:shadowing}
\end{figure}

In order to minimize the effect of time evolution as much as possible, we propose a paradigm in which the neuronal inputs and outputs are periodically measured in real time and the simulator parameters are updated during training. This method minimizes the neural fatigue caused by excessive stimulation time while addressing the sim-to-real problem. We refer to this learning paradigm as parameter shadowing (Fig.~\ref{fig:shadowing}).

\section{Implementations}
All experiments were conducted with the approval of local ethics committee of the University of Tokyo. This study was carried out in accordance with ``Guiding Principles for the Care and Use of Animals in the Field of Physiological Science'' published by the Japanese Physiological Society. The experimental protocol was approved by the Committee on the Ethics of Animal Experiments at the Research Center for Advanced Science and Technology, the University of Tokyo (Permit Number: RAC130106).

\subsection{Biological neuron cultures}
\subsubsection{Cell culture}
The protocol of preparing cell culture followed what was previously described \cite{bakkum2013tracking,muller2013sub,yada2017development}. In brief, we used the cerebral cortex of fetuses removed from Wistar rats on day 18 of gestation. After dissection and filtration of the cerebral cortex, the density was adjusted to approximately 38000 cells per 20 ${\mu}L$ using a cell counter. The dissociated cells were plated on high-density CMOS multi-electrode arrays (MEA) (MaxOne, MaxWell Biosystems) and fed a culture medium. The days in vitro of the culture sample used in this study was 42.

\subsubsection{Electrical interface of neuronal cultures}\label{neuron_io}
\begin{figure}
	\centering
	\includegraphics[width=0.8\linewidth]{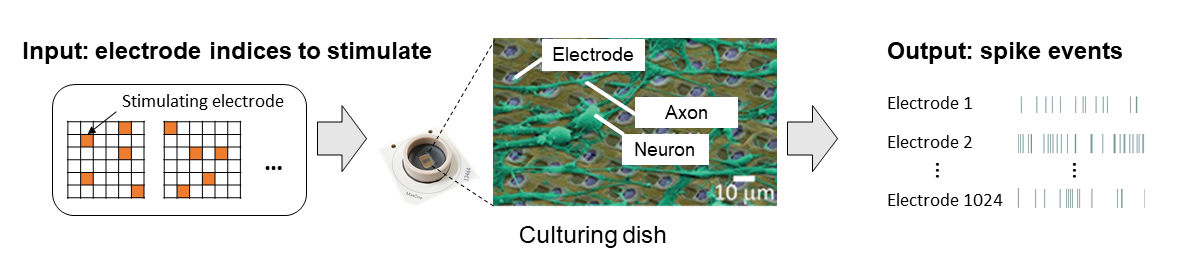}
	\caption{Electrical interface of neuronal cultures.}
	\label{fig:neuron_io}
\end{figure}

Fig.\ref{fig:neuron_io} shows the electrical interface of neurons. The CMOS MEA features 26400 platinum electrodes in a 3.85~$mm$ × 2.10~$mm$ area, wherein signals can be simultaneously recorded from up to 1024 arbitrary electrodes with a 20~$kHz$ sampling rate~\cite{ballini20141024}. In addition, up to 32 of the 1024 recording sites can be used to elicit neural spikes using stimulation electrodes placed adjacent to each recording electrode. The recording electrode is 9.3x5.4~${\mu}m$ in size and are placed at a pitch of 17.5 ${\mu}m$ (3265 electrodes/$mm^{2}$). The on-chip circuit amplified and band-pass filtered (100-3000 $Hz$) the obtained signals~\cite{bakkum2013tracking}.
In selecting electrodes for recording, spontaneous activity was first recorded for 20 seconds by scanning the entire electrode space.% for each grid containing no more than 1024 electrodes. 
Among the electrodes with a signal-to-noise ratio above a threshold, the top 1024 electrodes that recorded high firing rates were defined as the recording site.

Electrodes for stimulating neurons were selected among the recording sites that were estimated to be near somas of excitatory neurons. The selection was based on the shape and the amplitude of the waveform. Note that the neural activity of excitatory neurons shows a bipolar waveform with a longer duration and recovery time than that of inhibitory neurons~\cite{mita2019classification}. A pattern of simultaneous stimulation was prepared by randomly selecting five electrodes that were more than 100~${\mu}m$ away from each other. The electrical stimulus was biphasic voltage pulses with -200~$mV$, 500~${\mu}s$ pulse followed by 200~$mV$, 400~${\mu}s$ pulse. In order to avoid plasticity due to repetitive stimulation at fixed locations, 100 different patterns of stimulation were prepared for the experiment.

\subsection{Configuration of Animats}
\begin{figure}
	\centering
	\includegraphics[width=0.6\linewidth]{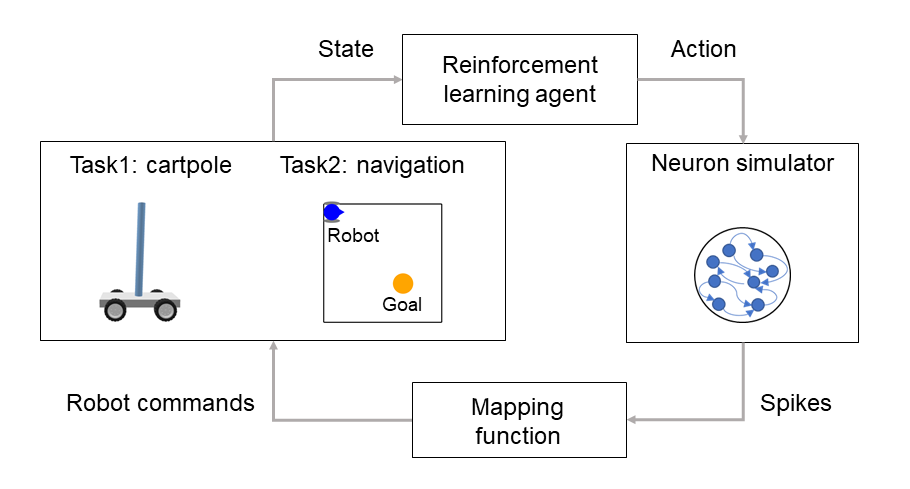}
	\caption{Overview of the configuration of Animats for two Animat conditions.}
	\label{fig:overview}
\end{figure}

Fig.~\ref{fig:overview} shows the pipeline for training and controlling Animats. We prepared two Animats for different tasks: balancing a pole and navigating to a goal in an arena. We call those the cartpole task and the navigation task. In both tasks, the controller observed the robot's state and output information on how to stimulate the simulated neurons. The output of the neurons drove Animats through a mapping function. For cartpole and navigation tasks, neural activities were mapped to a force to a cart and the orientation of a mobile robot, respectively.

\subsubsection{Neuron simulator}\label{neuron_sim}
As discussed in Section~\ref{neuron_model}, we prepared neuron simulators based on data recorded from cultured neurons. As neuron's responsiveness is known to be subject to the frequency of input signal~\cite{buzsaki2004neuronal, thomson2003interlaminar}, we designed the simulator to take discrete values of stimulus frequency as input and to output the number of spikes estimated to be evoked. Five frequencies were considered in this study (i.e., 5, 10, 20, 40, 80 $Hz$) within the reasonable range for neurons to operate~\cite{buzsaki2004neuronal}. 

\begin{figure}[ht]
	\centering
	\includegraphics[width=0.6\linewidth]{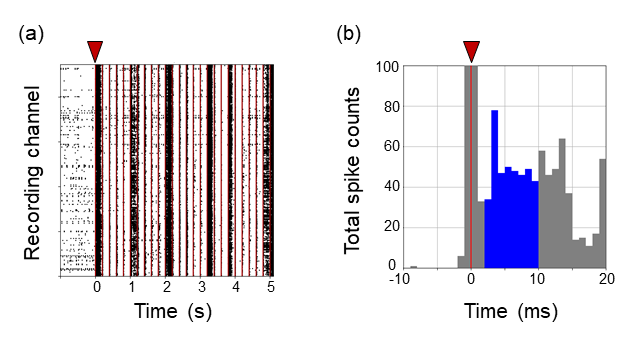}
	\caption{Recorded spike events for 5 Hz stimuli. (a) Raster plot of neural firing activity during electrical stimulation. The red lines indicate the timing of stimulation. (b) Spike-time histogram around after a stimulus (red triangle). Note that the artifact around the stimulation time (i.e., -1-1~$ms$) are out of the vertical range. Spike events between 2 $ms$ and 10 $ms$ post-stimulus were used in this study (blue bars).}
	\label{fig:raster_plot}
\end{figure}

\subsubsubsection{\textbf{Data recording}}
A data for each frequency condition was recorded as follows: first, a set of 100 stimulation patterns (see Section~\ref{neuron_io}) were presented to the cultured neurons in a random order with an inter-stimulate interval matching the frequency. The evoked neural responses were recorded during the stimuli (Fig.~\ref{fig:raster_plot} (a)). A neural response to a stimulus pattern was summarized as the total number of spikes between 2 $ms$ and 10 $ms$ after the stimulus, $S_{total}$ (Fig.~\ref{fig:raster_plot} (b)), and converted into a firing rate per channel, $x_f$, using the following equation:
\begin{equation}
  x_f = \frac{1}{TN} S_{total},
\end{equation}
where $T$ and $N$ represent the length of focused time window (i.e., 8 $ms$) and the number of channels, respectively. Note that the data immediately after stimuli  (i.e., 2 $ms$) were excluded to avoid counting electrical artifacts. Since we presented 100 stimulus patterns, 100 samples of $x_f$ were collected for each frequency condition. The recording of the five frequency conditions were conducted in a random order of conditions to eliminate order effects, and took about 120 $s$ in total.

\begin{figure}[ht]
	\centering
	\includegraphics[width=0.6\linewidth]{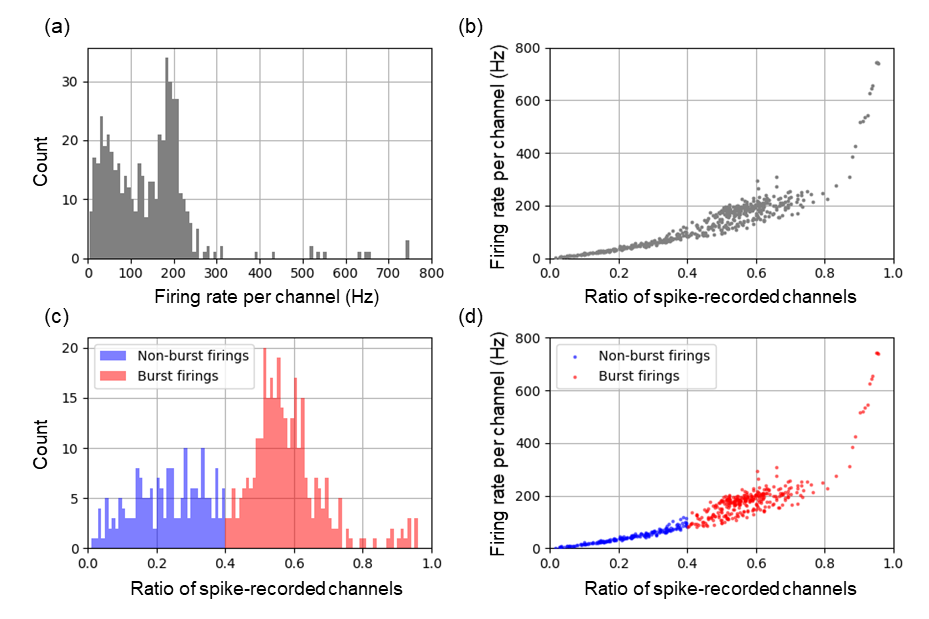}
	\caption{Extraction of non-burst activities. (a) Histogram of the firing rate recorded across five frequency conditions (5, 10, 20, 40, and 80 Hz). (b) Firing rate plotted against the ratio of channels that recorded at least one spike within the recording time window of 2 $ms$ and 10 $ms$ post-stimulus. (c) Thresholding of the spike-recorded channels based on Otsu's method to classify non-burst and burst firings. (d) The same plot as (b) with the labels of non-burst and burst firings.}
	\label{fig:non-burst_extraction}
\end{figure}

\subsubsubsection{\textbf{Data processing for preparing simulators}}
Neural responses shows burst firings, which is characterized by synchronized neural activities (Fig.~\ref{fig:non-burst_extraction} (a)). We focused on non-burst activity because burst firings unpredictably happen irrespective of stimulus frequencies. We classified the sampled data into burst and non-burst based on the ratio of channels that recorded spikes (Fig.~\ref{fig:non-burst_extraction} (b)). A data $x_f$ was classified as non-burst firings if the ratio of channels was sub-threshold (Fig.~\ref{fig:non-burst_extraction} (d)). Threshold was obtained by Otsu's method (Fig.~\ref{fig:non-burst_extraction} (c)) performed across five frequency conditions (i.e., 500 samples) at once. The non-burst data were summarized in a normalized histogram with 20 bins in a range between 0 to 120 Hz, which was determined experimentally. This histogram was defined as a neuron simulator by considering it as a probability distribution of neural activity for a given frequency stimulus.

\subsubsubsection{\textbf{Use of simulators during the training of policies}}
During training, an RL agent chose one of the five frequencies at every iteration and the simulator returned a value sampled based on the corresponding probability distribution. Since plasticity occurs in the biological neural network, the gap between neural simulators and real neurons increases during the training. For the simulator to reflect changes in actual neuronal properties, we recorded data every 10 minutes and updated the simulators while the training was carried out (i.e., parameter shadowing; see Section \ref{parameter_shadowing}).

%\subsubsection{Mapping function}
%We prepared several mapping functions that bridge the output of neural simulator (i.e., total spike counts) and the force added to the cartpole. 

%\begin{description}
%      \item[Linear mapping] the output of the cultured-neuron simulator was linearly mapped to the defined force range using the following equation:
%      \item[Percentile-based mapping of binary signals] the output of the cultured-neuron simulator was mapped to a force value according based on the value range below:
%      \item[Percentile-based mapping of multi signals] the output of the cultured-neuron simulator was mapped to a force value according based on the value range below:
%\end{description}

\subsubsection{Animat simulator for cartpole task}\label{Cartpole_sim}
An existing open-sourced simulator was used in this study \cite{githubGitHubMicrosoftcartpolepy}. The simulation parameters are shown in Table~\ref{tab:simulationparams}. The state of the cartpole was updated at every training step based on the input from the neuron simulator, $x$. 
%In addition, a small noise was added to the force at each iteration to simulate an uncertainty that may arise in a real-world environment. 
%so a policy cannot succeed by simply applying zero force.
To explore suitable mapping, we prepared several mapping functions that bridge the output of the neuron simulator (i.e., total spike counts) and the force added to the cartpole. 
\begin{description}
      % \item[Linear mapping] the output of the neuron simulator $x$ was linearly mapped to a force $f$ within the range of $-1-1$ using the following equation:
      % \begin{equation}
      % f=min \big(1, max  \big(-1,  \frac{1}{a-b} \big(2x - (a+b) \big) \big) \big), 
      % \end{equation}
      % where $a$ and $b$ represent the maximum and the minimum of non-burst firings collected by actual recording, respectively.
      \item[Percentile-based mapping for binary signals (1-threshold)] $x$ was mapped to $f$ added to value using the following equation:
      \begin{equation}
      f =\begin{cases}1 & if \quad x \geq  P_{50} \\-1 & if \quad x < P_{50}\end{cases}, 
      \end{equation}
      where $P_{50}$ represents the 50 percentile of non-burst firings collected across the five frequency conditions.
      \item[Percentile-based mapping for multiple signals (9-thresholds)] $x$ was mapped to $f$ added to value using the following equation:
      \begin{equation}
        f =\begin{cases}
                -1.0 & if \quad P_{0} \leq x \leq  P_{10} \\
                -0.8 & if \quad P_{10} < x \leq  P_{20} \\
                \quad \vdots \\
                0.8 & if \quad P_{80} < x \leq  P_{90} \\
                1.0 & if \quad P_{90} < x \leq  P_{100}
            \end{cases}.
      \end{equation}
\end{description}

\begin{table}[ht]
\caption{Parameters of the cartpole task}
\centering
\begin{tabular}{|r|l|}
\hline
Parameter     & Value (unit) \\ \hline
Gravity       & 9.8 ($m/s^{2}$)    \\ %\hline
Cart mass     & 3.1$\times10^{-1}$ ($kg$)    \\ %\hline
Pole mass     & 5.5$\times10^{-1}$ ($kg$)    \\ %\hline
Pole length   & 4.0$\times10^{-1}$ ($m$)     \\ %\hline
Track width   & 1.0 ($m$)      \\ %\hline
Step duration & 2.0$\times10^{-2}$ ($s$)     \\ %\hline
Force range   & -1.0-1.0 ($N$) \\ %\hline
Noise to the force & -2.0$\times10^{-2}$-2.0$\times10^{-2}$ ($N$)\\ \hline
\end{tabular}
\label{tab:simulationparams}
\end{table}

\subsubsection{Animat simulator for navigation task}\label{Mobile-robot_sim}
A mobile-robot simulator was prepared for the navigation task. The simulation parameters are shown in Table~\ref{tab:simulationparams_mobilerobot}. The state of the mobile robot was defined as the position of the robot and its orientation. At every training step, the robot updates the orientation, and moves forward by \textit{step duration}.

We prepared several mapping functions that bridge the output of the neuron simulator (i.e., total spike counts) and the value added to a mobile robot. 
\begin{description}
      % \item[Linear mapping] the output of the neuron simulator $x$ was linearly mapped to a modification of the robot's angle $\Delta \theta$ within the defined range of $-pi/10-pi/10$ using the following equation:
      % \begin{equation}
      % \Delta \theta=min \big(pi/10, max  \big(-pi/10,  \frac{1}{a-b} \big(2x - (a+b) \big) \big) \big), 
      % \end{equation}
      % where $a$ and $b$ represent the maximum and the minimum of non-burst firings collected by actual recording, respectively.
      \item[Percentile-based mapping for binary signals (1-threshold mapping)] $x$ was mapped to $\Delta \theta$ using the following equation:
      \begin{equation}
      \Delta \theta =\begin{cases}pi/10 & if \quad x \geq  P_{50} \\-pi/10 & if \quad x < P_{50}\end{cases}, 
      \end{equation}
      where $P_{50}$ represents the 50 percentile of non-burst firings collected across the five frequency conditions.
      \item[Percentile-based mapping for multiple signals (9-thresholds mapping)] $x$ was mapped to $\Delta \theta$ using the following equation:
      \begin{equation}
        \Delta \theta = pi/10 * \begin{cases}
                -1.0 & if \quad P_{0} \leq x \leq  P_{10} \\
                -0.8 & if \quad P_{10} < x \leq  P_{20} \\
                \quad \vdots \\
                0.8 & if \quad P_{80} < x \leq  P_{90} \\
                1.0 & if \quad P_{90} < x \leq  P_{100}
            \end{cases}.
      \end{equation}
\end{description}

\begin{table}[ht]
\caption{Parameters of the navigation task}
\centering
\begin{tabular}{|r|l|}
\hline
Parameter     & Value (unit) \\ \hline
Arena height and width       & 1.0 ($m$)    \\ %\hline
Step length per iteration     & 0.025 ($m$)    \\ %\hline
Angle range     &  -pi-pi($rad$)    \\ %\hline
Angle modification range   & -pi/10-pi/10($rad$)    \\ \hline
\end{tabular}
\label{tab:simulationparams_mobilerobot}
\end{table}

\subsubsection{Training of policy using reinforcement learning}
We used Microsoft Bonsai platform for training an RL agent \cite{microsoftProjectBonsai,zhao2021reinforcement}. Bonsai is characterized by its natural-language-like programming interface and capabilities to automatically optimize deep neural networks. In this study, we used Bonsai to focus on the design of Animats that aim for context-dependent behavior rather than designing network architectures and learning algorithms for RL.

For the training of the cartpole task, an agent observed a set of parameters that represent the states of the cartpole: cart position, cart velocity, pole angle (radian), pole angular velocity (radian/s). For the training of the navigation task, an agent observed the position and the orientation of the robot. Refer to Section~\ref{experiments} for details including other learning parameters and reward functions.

\section{Experiments}\label{experiments}
Prior to the learning experiments using the neuron simulator, actual neural activity was measured over 2 hours to visualize changes in neural responses across frequencies and times. We also solved the cartpole and navigation tasks without the neuron simulator to confirm the baseline performance of the policies learned by Bonsai. Finally, we solved these tasks using a neuron simulator to see if the proposed design would enable goal-oriented control of Animat.

\subsection{Recording of neural responses}\label{neural_recording}
\begin{figure}[ht]
	\centering
	\includegraphics[width=0.6\linewidth]{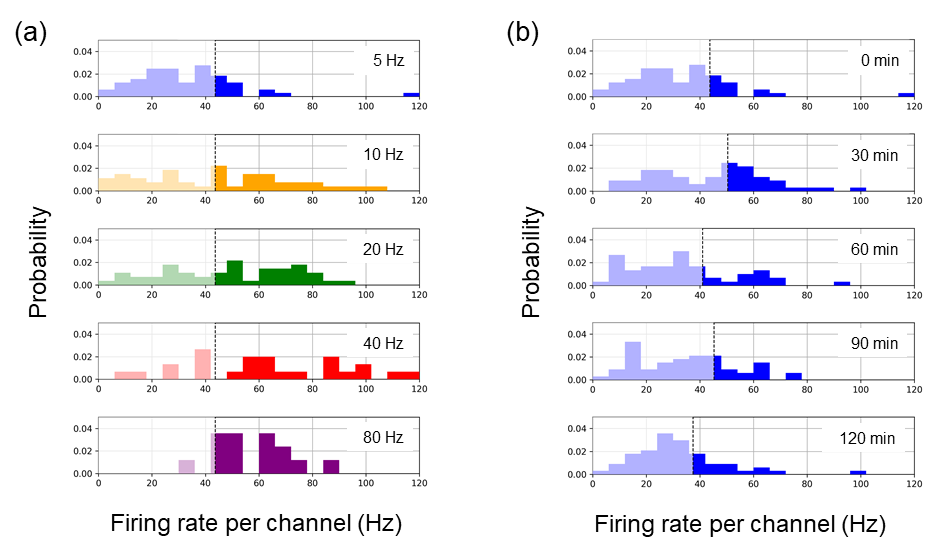}
	\caption{Differences in response intensity for different stimulus conditions. Dot lines indicate the 50 percentile of non-burst firings collected across the five frequency conditions. (a) Comparison of the measured response intensity for different stimulus frequencies at the first recording in a day. (b) The temporal evolution of the response intensity to a stimulation frequency of 5 Hz on the same day.}
	\label{fig:histogram_comparison}
\end{figure}

Fig.~\ref{fig:histogram_comparison} (a) shows the comparison of the measured response intensity for different stimulus frequencies. The distribution of response intensity showed a different trend depending on the stimulus frequency. In particular, as the frequency increased, small response intensities were less likely to be produced and the distribution seemed to shift to larger values.
Fig.~\ref{fig:histogram_comparison} (b) shows the temporal evolution of the response intensity to a stimulation frequency of 5 Hz. It appeared that the high-intensity responses became less apparent with time, possibly due to neural fatigue. These results indicate that neural activity depends on the stimulus frequency and also changes over time.

\subsection{Training without the neuron simulators}
\subsubsection{Cartpole task}
\begin{figure}[ht]
	\centering
	\includegraphics[width=0.6\linewidth]{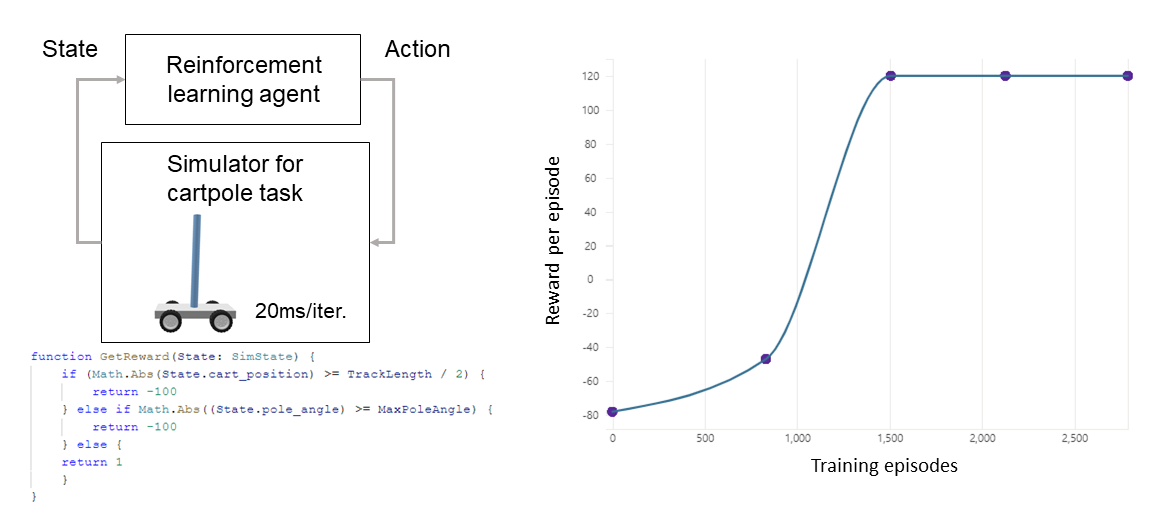}
	\caption{Training results without the cultured-neuron simulator (cartpole task).}
	\label{fig:without_cultured-neuron_simulator_simple}
\end{figure}

We verified that Bonsai is capable of solving the cartpole task in a configuration that excludes the neuron simulator. The detailed parameters for the training are shown in Table~\ref{tab:parameters_wo_cultured-neuron_simple}. The rewards are designed to motivate agents to balance the pole for as long as possible while avoiding falling poles and overrunning the cart. Specifically, at every training iteration, the agent was rewarded with the value of $1$ unless the pole tilts over 12 ($rad$) or the cart overruns the track. If one of the two cases happens, the agent was rewarded with the value of $-100$ and the episode was terminated. The training continued until the number of training iterations reached a preset number (see Table~\ref{tab:parameters_wo_cultured-neuron_simple}). Fig.~\ref{fig:without_cultured-neuron_simulator_simple} shows the result. The robot gradually learned to balance the pole during an episode of 120 iterations, showing the validity of the use of Bonsai.

\begin{table}[ht]
\caption{Parameters for the cartpole task without the neuron simulator}
\centering
\begin{tabular}{|c|c|}
\hline
Parameter & Value \\ \hline
Maximum iterations per episode & 120 \\ %\hline
Action (force to the cart)& -1.0 or 1.0 \\ %\hline
Algorithm   & Soft Actor-Critic (SAC) \\
Pole angle at the beginning of episodes   & 0 ($rad$) to the vertical direction\\
Limit of iterations to terminate training & 800000 \\
\hline
\end{tabular}
\label{tab:parameters_wo_cultured-neuron_simple}
\end{table}

\subsubsection{Navigation task}
\begin{figure}[ht]
	\centering
	\includegraphics[width=0.6\linewidth]{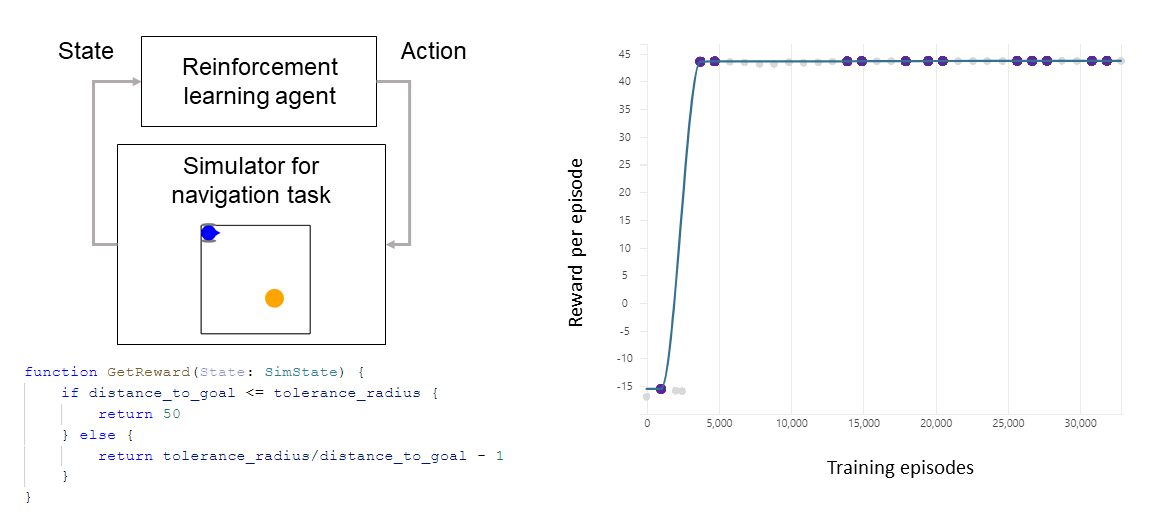}
	\caption{Training results without the cultured-neuron simulator (navigation task).}
	\label{fig:without_cultured-neuron_simulator_maze}
\end{figure}

We verified that Bonsai is capable of solving the navigation task in a configuration that excludes the neuron simulator. The detailed parameters for the training is shown in Table~\ref{tab:parameters_wo_cultured-neuron_maze}. The rewards are designed to motivate agents to reach the goal position as soon as possible based on the following equation:
      \begin{equation}
        Reward(d) = \begin{cases}
                50 & if \quad d \leq R_{goal} \\
                R_{goal}/d -1 & \quad if \quad d > r
            \end{cases},
      \end{equation}
where $d$ and $R_{goal}$ represent the distance to the goal from the robot and the threshold at which the robot is defined to have reached the goal, respectively. The training continued until the number of training iterations reached a preset number (see Table~\ref{tab:parameters_wo_cultured-neuron_maze}). Fig.~\ref{fig:without_cultured-neuron_simulator_maze} shows the result. The agent gradually learned the policy to reach the goal, suggesting the validity of the use of Bonsai.

\begin{table}[ht]
\caption{Parameters for the navigation task without the neuron simulator}
\centering
\begin{tabular}{|c|c|}
\hline
Parameter & Value \\ \hline
Maximum iterations per episode & 20 \\ %\hline
Action ($\Delta \theta$)& -pi/10 - pi/10 ($rad$) \\ %\hline
Algorithm   & Soft Actor-Critic (SAC) \\
Robot position and orientation & The top-left corner facing right (Fig.~\ref{fig:without_cultured-neuron_simulator_maze}) \\
at the beginning of episodes  & \\
Goal position & (0.7, 0.7) from the top-left corner \\
& (the orange area in Fig.~\ref{fig:without_cultured-neuron_simulator_maze}) \\
$R_{goal}$ & 0.1 \\
Limit of iterations to terminate training & 500000 \\
\hline
\end{tabular}
\label{tab:parameters_wo_cultured-neuron_maze}
\end{table}

%\subsubsection{Comparison between different RL algorithms}
%\subsubsection{Comparison with the half-embodied condition}

\subsection{Training Animats}
We trained to solve the tasks in a configuration including the neuron simulators as shown in Fig.~\ref{fig:overview}. We used 14 sets of neuron simulators that were recorded every 10 minutes for 140 minutes in section~\ref{neuron_sim}. For parameter shadowing, training started using the neuron simulators at the beginning of the recording (i.e., Fig.~\ref{fig:histogram_comparison} (a)) and updated them every 10 minutes to match the interval of recording. 
The training continued until the number of training iterations reached a preset number (see Table~\ref{tab:parameters_wo_cultured-neuron_simple} and \ref{tab:parameters_wo_cultured-neuron_maze}).
The detailed parameters for the training were the same as Table~\ref{tab:parameters_wo_cultured-neuron_simple} and \ref{tab:parameters_wo_cultured-neuron_maze} except that the action was defines as one of the five indices corresponding to the five stimulus frequencies (see Fig.~\ref{fig:histogram_comparison} (a)).

To compare the performances of Animats across conditions, ten policies were trained for each mapping function (see section~\ref{Cartpole_sim} and section~\ref{Mobile-robot_sim}). The performances were evaluated based on the reward values acquired from the environment. To check the progress of training and the effectiveness of mapping functions, the following comparisons were made for the both tasks:
\begin{enumerate}
\item Performances at the beginning and after the training under the 1-threshold condition
\item Performances after the training under a control condition and under the 1-threshold condition
\item Performances after the training under 1-threshold condition and under the 9-thresholds condition,
\end{enumerate}
where the control condition indicates random mapping between the output of Bonsai policy and the selection of neuron simulator. For the comparison $1$, performances of 30 episodes were evaluated using the neuron simulators based on measured data at the start of recording (i.e., Fig.~\ref{fig:histogram_comparison} (a)). For the other comparisons, performances of 100 episodes were evaluated using the neuron simulators based on measured data 140 min after the start of recording.

To test the effectiveness of parameter shadowing, we compared the performance of Animats trained with and without parameter shadowing using the 1-threshold mapping. In the condition without Parameter shadowing, the policy was trained without updating the neuron simulators. For the other comparisons, performances of 100 episodes were evaluated using the neuron simulators based on measured data 140 min after the start of recording.

\subsubsection{Cartpole task}
\begin{figure}[ht]
	\centering
	\includegraphics[width=0.6\linewidth]{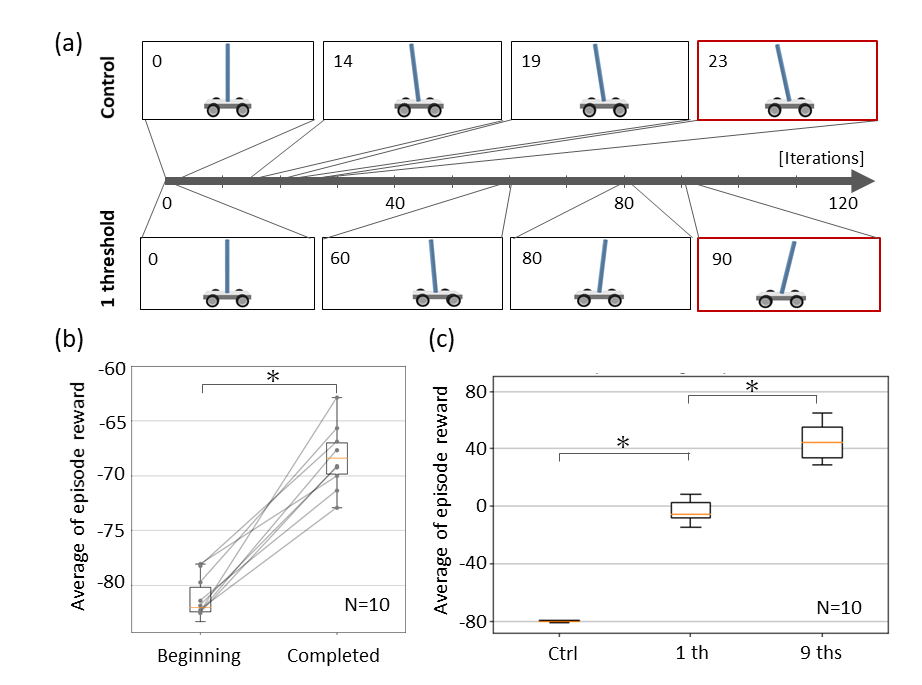}
	\caption{Result of training Animats for the cartople task. (a) Examples of the Animat's behavior under the control and the 1-threshold conditions. The frames with red rectangles indicate the termination of an episode because the pole angles exceeded tolerance. (b) Comparison of performances at the beginning and after the training under the 1-threshold condition. (c) Comparisons of performances after the training. Asterisks indicate significant differences ($p<0.01$, Mann-Whitney U test).}
	\label{fig:result_cartpole}
\end{figure}

Since all training was completed within 60 minutes, a maximum of six pairs of neuron simulators were used during the training. Fig.~\ref{fig:result_cartpole} (a) shows examples of the Animat's behavior after the training. Policies trained with the 1-threshold mapping balanced the poles for longer iterations than those trained under the control condition. Fig.~\ref{fig:result_cartpole} (b) shows the comparison of performances at the beginning and after the training under the 1-threshold condition. The performance increased significantly after the training ($p=1.82\times10^{-4}$, Mann-Whitney U test), suggesting that RL improved the performance of the Animat. Fig.~\ref{fig:result_cartpole} (c) shows the comparisons of performances after the training. Policies trained with the 1-threshold condition showed greater performance than those trained under the control condition ($p=1.83\times10^{-4}$, Mann-Whitney U test), indicating the effectiveness of the threshold-based mapping for the training. In addition, policies trained with the 9-thresholds condition showed greater performance than those trained with the 1-threshold condition ($p=1.83\times10^{-4}$, Mann-Whitney U test), suggesting the effectiveness of higher resolution for training.

\subsubsection{Navigation task}
\begin{figure}[ht]
	\centering
	\includegraphics[width=0.6\linewidth]{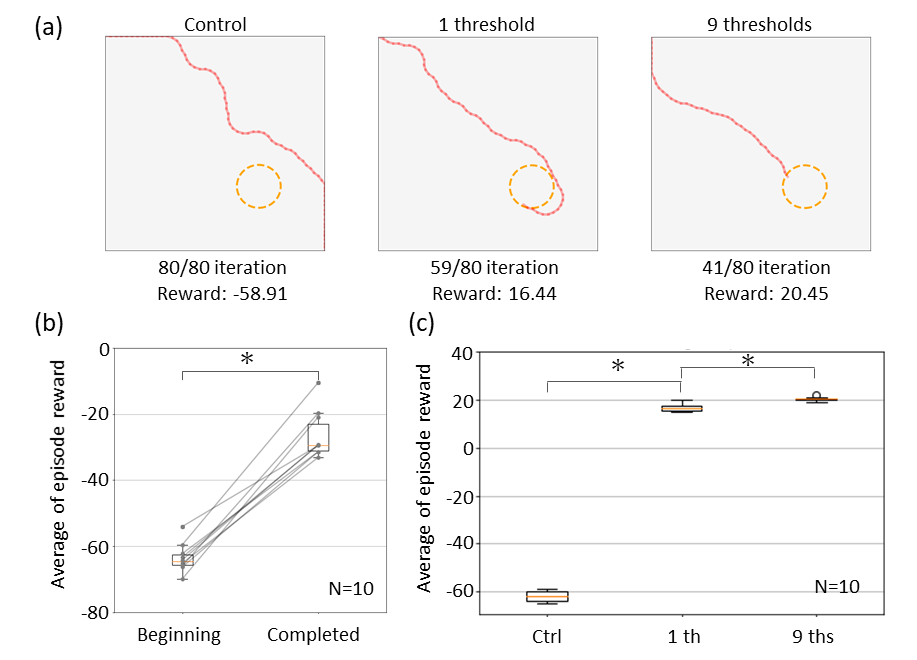}
	\caption{Result of training Animats for the navigation task. (a) Examples of the Animat's behavior under the three conditions. Orange dashed circles indicate the goal area. (b) Comparison of performances at the beginning and after the training under the 1-threshold condition. (c) Comparisons of performances after the training. Asterisks indicate significant differences ($p<0.01$, Mann-Whitney U test). }
	\label{fig:result_wheel}
\end{figure}

Since all training was completed within 60 minutes, a maximum of six pairs of neuron simulators were used during the training. Fig.~\ref{fig:result_wheel} (a) shows examples of the Animat's behavior after the training. The Animat failed to reach the goal under the control condition, and reached the goal faster under the 9-thresholds condition than the 1-threshold condition. Fig.~\ref{fig:result_wheel} (b) shows the comparison of performances at the beginning and after the training under the 1-threshold condition. The performance increased significantly after the training ($p=1.83\times10^{-4}$, Mann-Whitney U test), suggesting that RL improved the performance of the Animat. Fig.~\ref{fig:result_wheel} (c) shows the comparisons of performances after the training. Policies trained with the 1-threshold condition showed greater performance than those trained under the control condition ($p=1.83\times10^{-4}$, Mann-Whitney U test), indicating the effectiveness of the threshold-based mapping for the training. In addition, policies trained with the 9-thresholds condition showed greater performance than those trained with the 1-threshold condition ($p=3.30\times10^{-4}$, Mann-Whitney U test), suggesting the effectiveness of higher resolution for training.

\subsubsection{Effectiveness of parameter shadowing}
\begin{figure}[ht]
	\centering
	\includegraphics[width=0.6\linewidth]{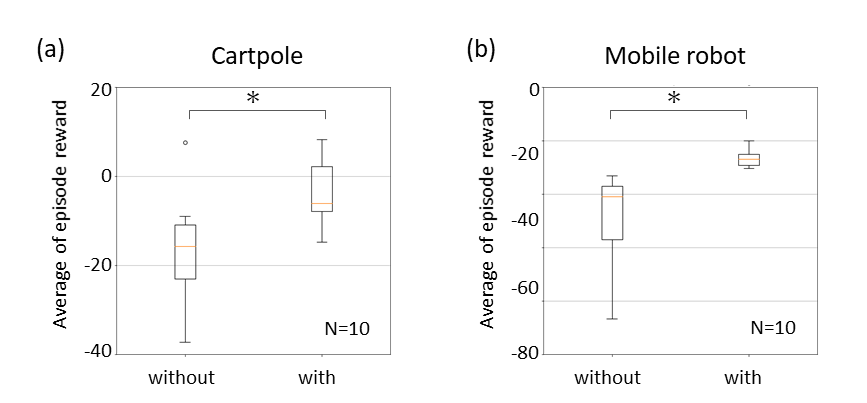}
	\caption{ Comparison of performances with and without parameter shadowing. (a) Cartpole task. (b) Navigation task. }
	\label{fig:parameter_shadowing}
\end{figure}

Fig.~\ref{fig:parameter_shadowing} shows a comparison of the performances with and without parameter shadowing. In both cartpole and navigation tasks, policies trained with parameter shadowing performed better than those trained without parameter shadowing ($p=7.28\times10^{-3}$ and $p=1.83\times10^{-4}$ for cartpole and navigation task, respectively. Mann-Whitney U test). These results suggest the effectiveness of parameter shadowing for training Animats.

\section{Discussion}
In this paper, we aimed to acquire context-dependent behaviors for Animats. To this end, we proposed the use of RL and a neuron simulator based on measured data, as well as updating the simulator during the training of policies. We implemented Animat according to these design principles and showed in multiple scenarios that Animat autonomously acquires context-dependent behaviors through RL.

Rewards obtained from the behavior of Animats increased as the training progressed (Fig.~\ref{fig:result_cartpole} (b) and Fig.~\ref{fig:result_wheel} (b)). Since rewards are higher in proportion to the achievement of the task objective, the policies through RL have contributed to the context-dependent behaviors of Animats. To the best of the authors' knowledge, this study is the first to show context-dependent behavior can be acquired in Animat.

The firing rate of neurons depended on the stimulus frequency (Fig.~\ref{fig:histogram_comparison}). This result is consistent with the results by Eytan et al. \cite{eytan2003selective}, who examined the stimulus frequency dependence of distributed cultured neurons. Thus, the neuron simulator employed in this study could reflect the physiological characteristics of biological neurons. 

The conversion scheme between neural activity and robot motion commands based on a percentile index may have contributed to improved control performance by reducing the influence of the stochastic behavior of neurons on the motion commands (Fig.~\ref{fig:result_cartpole} (c) and Fig.~\ref{fig:result_wheel} (c)). Under the 1-threshold condition, given an undesired robot motion command (e.g., moving the cart to the left when it should move to the right), the force shift applied to the cart is $2$ [N]. On the other hand, under the 9-thresholds condition, when an undesirable robot motion command is given, there are 9 different force deviations from $0.2$-$2$ [N], and the expected value of the force deviation is smaller than under the 1-threshold condition. This result indicates the effectiveness of preparing a mapping function that minimizes the error in the motion command value due to the stochastic behavior of neurons.

Although this experiment suggests the feasibility of neuronal-mediated goal-directed robot behavior, the learning was conducted on a simulator. Therefore, it is not clear whether the learned strategies will work in a closed loop incorporating actual cultured neurons. In addition, the general usefulness of the conversion scheme between neural activity and robot motor commands based on percentile indices has not been demonstrated, toward the development of a generic learning framework for goal-oriented behavior in Animat. It is expected that these research issues will be resolved.

\section{Conclusion}
We have demonstrated that our proposed design principles can be applied to the control of Animat in multiple scenarios. Although there is a history of research attempting to control Animat, previous work of ``Braitenberg vehicle-type'' robots required the experimenter to design in advance the relationship between Animat's sensor information and the actions to be taken by Animat in light of the task objectives. In contrast, our proposed RL-based training enables Animat to autonomously learn policies for selecting the optimal action depending on the context in which it is placed. We believe that the proposed design principles provide a basis for investigating how neural information processing can be applied to robot control.

\bibliographystyle{unsrtnat}
\bibliography{references}  %%% Uncomment this line and comment out the ``thebibliography'' section below to use the external .bib file (using bibtex) .

\end{document}